\documentclass{article}

    \PassOptionsToPackage{numbers, compress}{natbib}
\usepackage[preprint, final]{neurips_2026}
\usepackage{graphicx}
\usepackage[utf8]{inputenc} % allow utf-8 input
\usepackage[T1]{fontenc}    % use 8-bit T1 fonts
\usepackage{hyperref}       % hyperlinks
\usepackage{url}            % simple URL typesetting
\usepackage{booktabs}       % professional-quality tables
\usepackage{amsfonts}       % blackboard math symbols
\usepackage{nicefrac}       % compact symbols for 1/2, etc.
\usepackage{microtype}      % microtypography
\usepackage{xcolor}         % colors
\usepackage{amsmath}

\title{WorldFly: A World-Model-Based Vision-Language-Action Model for UAV Navigation}

\author{
Shengtao Zheng$^{1*}$ \quad
Kai Li$^{1*}$ \quad
Weichen Zhang$^{1}$ \quad
Yu Meng$^{1}$ \\
Chen Gao$^{2\dagger}$ \quad
Xinlei Chen$^{1\dagger}$ \quad
Yong Li$^{2}$ \quad
Xiao-Ping Zhang$^{1}$ \\
$^{1}$Tsinghua Shenzhen International Graduate School \\
$^{2}$BNRist, Tsinghua University \\
\texttt{\{st\_zheng24, likai24, zhangwc23, mengy24\}@mails.tsinghua.edu.cn} \\
\texttt{chgao96@tsinghua.edu.cn, \quad chen.xinlei@sz.tsinghua.edu.cn} \\
\texttt{liyong07@tsinghua.edu.cn, \quad xpzhang@ieee.org}
}

\begin{document}

\maketitle
\footnotetext[1]{Equal contribution.}
\footnotetext[2]{Corresponding authors.}

\begin{abstract}
End-to-end Vision-Language-Action (VLA) models have shown promise in UAV navigation. However, existing approaches typically rely on historical observations to directly predict actions, often struggling in dense urban environments where severe occlusions and sharp turns result in drastic viewpoint transitions. We argue that the ability to ``imagine'' future states—inherent in World Models—is critical for robust decision-making under such partial observability. To address this, we construct a challenging Urban Canyon Traversal Benchmark, specifically designed to evaluate spatial understanding in scenarios characterized by severe occlusions and drastic viewpoint transitions. To this end, we propose WorldFly, a novel world-model-based VLA framework that employs a dual-branch coupled flow matching mechanism to jointly generate future video predictions and navigation actions, thereby explicitly guiding the agent's policy via spatial imagination. Extensive evaluations on our benchmark demonstrate that WorldFly outperforms other baselines, particularly in unseen environments, validating the effectiveness of integrating world models into embodied aerial agents. 
\end{abstract}

% \begin{figure}
%   \centering
%     \includegraphics[width=\columnwidth]{image/p1.pdf}
%     \caption{
%         WorldFly performs imagination and action decision-making simultaneously in urban environments, conditioned on human language instructions and historical observations.
%     }
%     \label{icml-historical}
% \end{figure}

\section{Introduction}
\label{sec:Introduction}
In urban low-altitude environments, ground mobile robots are often constrained by complex factors such as traffic conditions, cluttered infrastructure, and limited fields of view. As a result, they typically require sophisticated behavioral strategies and extensive prior mapping to achieve effective navigation. In contrast, Unmanned Aerial Vehicles (UAVs) benefit from low-altitude flight capabilities that allow them to bypass many ground-level constraints, enabling more flexible and efficient navigation in urban spaces and demonstrating superior accessibility.

Recent works, such as OpenFly~\cite{openfly} and UAV-Flow~\cite{uav-flow}, have begun to explore UAV navigation driven directly by human natural language instructions without expert intervention. Under the VLA paradigm, these methods aim to end-to-end generate navigation actions based on first-person-view (FPV) visual observations and language instructions. Despite promising progress, existing UAV VLA approaches, such as OpenVLA~\cite{openvla} and PI-0-UAV~\cite{uav-flow}, primarily rely on historical and current observations to predict future actions. They lack explicit estimation of future scene states, which significantly limits their ability to perform long-horizon action prediction in previously unseen environments. 
Moreover, when UAVs execute large-magnitude motions, observations across time steps can exhibit substantial semantic variation, making it difficult to temporally align visual observations with language instructions.

Meanwhile, in domains such as autonomous driving and video generation, world models have demonstrated strong capabilities in generating first-person-view videos conditioned on language descriptions, showcasing powerful alignment between text and visual modalities. Furthermore, interactive world models such as NWM~\cite{bar2025navigation}, Matrix-Games~\cite{he2025matrix}, and YUME~\cite{mao2025yume} have shown an impressive ability to model visual dynamics and action representations jointly. These advances motivate our core insight: by leveraging a world model to generate future visual trajectories conditioned on UAV instructions, we can extract motion cues from imagined videos to predict future actions and thereby enable more robust UAV navigation.

However, incorporating world models into UAV VLA frameworks presents several key challenges:
% 1) \textbf{Limited long-range generation capability}. Existing world models are primarily designed to generate short-horizon videos with small motion magnitudes. Their ability to model long-range, large-scale UAV movements in open urban environments remains severely limited.
% 2) \textbf{High computational cost}. World models typically require multi-step denoising processes to produce final predictions. A naive serial training pipeline—where the world model is trained first and then used to augment an action policy—introduces significant training and inference overhead.
% 3) \textbf{Error accumulation in long-horizon reasoning}. In open-space navigation tasks, long-range action prediction errors can accumulate rapidly, causing the UAV to deviate substantially from the intended instruction-aligned trajectory.
1) Existing UAV navigation datasets often overlook the extreme viewpoint variations and complex spatial constraints typical of urban "canyon" environments. Most benchmarks focus on tasks in low-clutter settings, failing to evaluate a model’s ability to handle instructions under drastic perspective shifts.
2) Current UAV VLA frameworks often treat visual perception and action prediction as sequential or loosely coupled modules. These models typically rely on reactive mapping of historical observations to current actions, lacking an integrated mechanism to jointly optimize future scene "imagination" with control policy learning.
3) Purely reactive VLA models struggle with "short-sightedness" in unfamiliar environments. Without an explicit estimation of future states or a world model to provide dense motion priors, these models cannot maintain temporal consistency, leading to failure when language instructions require navigating across long distances with significant semantic variation.

To address these challenges, we propose \textbf{WorldFly}, a unified world-model-enhanced VLA framework for UAV navigation. First, to tackle the first limitation, we introduce a new benchmark named Urban Canyon Traversal, which is characterized by large viewpoint variations and extended navigation trajectories. This benchmark is specifically designed to stress-test the generative capability of world models in complex urban environments.
Second, to improve architectural coupling between imagination and policy learning, we propose a dual-branch coupled architecture that enables the model to simultaneously generate future visual representations and navigation actions. This design jointly optimizes world modeling and action prediction within a single framework, avoiding a separately trained serial world-model-then-policy pipeline.
% Finally, to mitigate error accumulation in long-horizon navigation, we introduce a video-aligned temporal voting mechanism. By aggregating weighted cached action sequences that are temporally aligned with the current step, this mechanism suppresses abrupt action changes and significantly improves navigation stability and robustness in long-range VLA tasks.
Finally, we validate that our joint modeling of future video and action trajectories significantly enhances navigation robustness across diverse difficulty levels. Experimental results show that WorldFly outperforms existing baselines by utilizing imagined scenes as predictive guidance, successfully reducing the "short-sighted" failures typical of purely reactive VLA models.
\begin{figure*}[t]
  \centering
  \includegraphics[width=\textwidth]{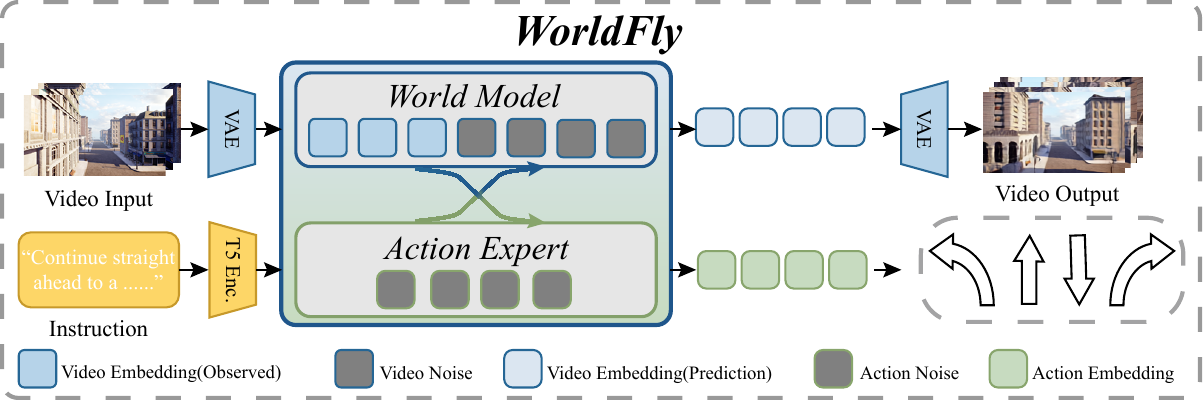}
  \caption{
   The overall architecture of our work. A dual-branch coupled flow matching mechanism is used to jointly generate future video predictions and navigation actions, integrating world modeling and policy learning in a unified Vision-Language-Action framework.
   % We further introduce an iterative multi-step rollout mechanism in the predicted visual latent space of the world model, where candidate actions from both current and past rollouts are temporally aligned and aggregated through an exponentially decayed voting scheme for robust execution.
  }
  \label{fig:worldfly}
\end{figure*}
Our contributions are summarized as follows:
\begin{itemize}
  \item We introduce a novel and challenging benchmark, \textbf{Urban Canyon Traversal}, designed to evaluate instruction-following capabilities of VLA models under significant viewpoint variations and long-horizon navigation requirements. The benchmark quantitatively measures UAV navigation performance in both seen and unseen urban environments.
  \item We propose \textbf{WorldFly}, the first framework to integrate world models into UAV VLA tasks. WorldFly employs a \textbf{Dual-Branch Coupled Architecture} that jointly optimizes future scene imagination and action generation, enabling future scene imagination while simultaneously generating navigation actions.
  % \item We further introduce a \textbf{Video-Aligned Temporal Voting} mechanism to enhance execution stability. By aggregating cached historical predictions via an exponentially decayed weighted voting scheme, our approach effectively combines future imagination with historical planning.
  \item Experiments show that \textbf{WorldFly} outperforms baseline models on the \textbf{Urban Canyon Traversal} benchmark, while jointly achieving future video generation and navigation action prediction within a unified framework.
\end{itemize}

\section{Related Work}

\subsection{UAV Navigation VLA}
UAV Vision-Language-Action (VLA) models generate future navigation actions by processing UAV visual observations and human language instructions. Existing works primarily adopt two paradigms within the VLA domain: the autoregressive action prediction architecture based on OpenVLA~\cite{openvla} (e.g., OpenFly~\cite{openfly}, RaceVLA~\cite{racevla}, OpenVLA-UAV~\cite{uav-flow}) and the Flow Matching architecture~\cite{flowmatching} based on $\pi_0$~\cite{pi0} (e.g., Pi-0-UAV~\cite{uav-flow}).

RaceVLA~\cite{racevla} utilizes 4-DoF velocity vectors as UAV control signals. It constructs a UAV motion dataset in real-world environments and builds navigation tasks for racing scenarios based on the OpenVLA model. OpenFly~\cite{openfly}, fine-tuned from OpenVLA, proposes a keyframe-aware navigation model. It extracts critical visual features during flight via an adaptive frame-level sampling mechanism and employs visual token fusion to effectively reduce information redundancy between adjacent frames while preserving historical visual memory. UAV-Flow~\cite{uav-flow} introduces a fine-grained, trajectory-guided instruction navigation benchmark. It trains OpenVLA-UAV and Pi-0-UAV to focus on understanding low-level flight semantics, thereby evaluating the UAV's capabilities regarding spatial perception and motion correlation.

However, these methods typically map historical observations directly to actions, lacking the capability to explicitly predict or ``imagine'' future environmental states. This limits their robustness in complex urban scenarios where spatial foresight is crucial for handling severe occlusions and drastic viewpoint transitions.

\subsection{World Model VLA}
Currently, VLA models based on World Models predominantly emerge in the field of robotic manipulation. World Model VLAs construct a deep integration of visual understanding, action generation, and environmental simulation, incorporating the spatial reasoning capabilities of world models into VLA architectures to enhance adaptability in complex scenes.

Genie-Envisioner~\cite{genie-envisioner} comprises two structural components: GE-Base, a large-scale instruction-conditioned video diffusion model trained on the AgiBot World dataset to predict future manipulation scenes; and GE-Act, an action head that maps latent visual feature representations into executable action trajectories via a lightweight flow matching decoder. VideoVLA~\cite{videovla} employs the VAE encoder from CogVideoX for video prediction, utilizing a unified vision-action structure and a Flow Matching architecture to simultaneously predict future video frames and generate actions. UVA~\cite{uva} focuses on the co-optimization of video generation and action prediction. It proposes a unified video-action latent feature representation alongside a decoupled diffusion decoding architecture. WorldVLA~\cite{worldvla} integrates the VLA model and the world model into a single autoregressive framework, achieving unified understanding and generation of both actions and images. MinD~\cite{mind} addresses the computational inefficiency of video generation models in real-time robotic control by proposing a dual-system diffusion architecture that decouples low-frequency visual imagination from high-frequency action generation.

Despite these advancements, existing World Model-based approaches remain largely confined to robotic manipulation within static, small-scale environments (e.g., tabletops). Applying them to UAV navigation presents unique challenges, primarily the need to handle large-scale urban topologies with drastic viewpoint changes and the stricter real-time inference requirements for aerial safety. WorldFly bridges this gap by extending the World Model VLA paradigm to the challenging domain of embodied aerial navigation.

\begin{figure*}[t]
  \centering
  \includegraphics[width=\textwidth]{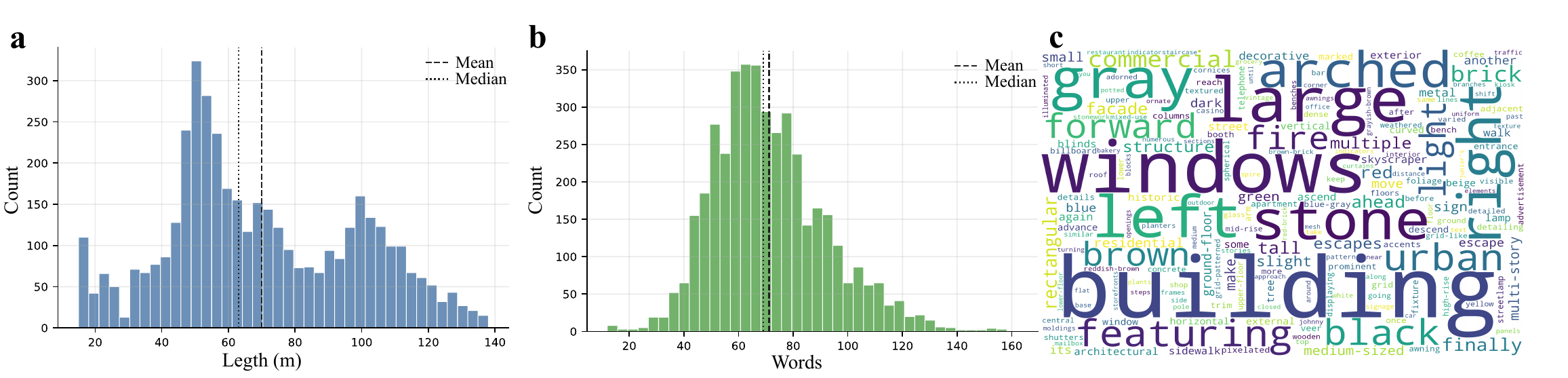}
  \caption{
    a. Trajectory length distribution. b. Word count distribution for the generated instructions. c. Word cloud of the text descriptions in the dataset.
  }
  \label{fig:worldfly}
\end{figure*}

\section{Data Collection and Benchmark Construction}

To evaluate the enhancement provided by World Models to UAV navigation capabilities, it is essential to design a benchmark characterized by complex variations in spatial features. In low-altitude urban scenarios, we observe that when UAVs navigate between buildings—particularly when executing sharp turns at street intersections—the visual features in the First-Person View (FPV) undergo drastic transitions. Consequently, we designed the Urban Canyon Traversal Benchmark specifically to assess the UAV's capability to execute tasks under these challenging conditions.

\paragraph{Benchmark Construction.}
Leveraging the OpenFly simulation toolchain, we utilized skeleton graphs to extract intersection nodes. To generate expert demonstrations, we identified 63 intersection waypoints and employed the A* algorithm to act as an expert policy guiding the UAV. To enhance generalization and ensure the retention of longer trajectories, target waypoints were randomly sampled near the ends of streets. The dataset was further post-processed by filtering out short-range and invalid trajectories.
We collected over 4,000 navigation trajectories across AirSim urban maps. Among them, 100 trajectories were selected to form the TEST-EASY evaluation benchmark, while the remaining trajectories were used for training. To assess generalization to unseen environments, we further collected 100 navigation trajectories at 14 newly introduced intersections, which constitute the TEST-HARD evaluation benchmark.
\paragraph{Instruction Generation.}
Navigation instructions are generated automatically using an LLM-based instruction constructor built on the OpenFly toolchain. Concretely, we convert expert routes and landmark cues into textual navigation descriptions using Qwen3-VL, producing instruction-following trajectories that retain the large-turn and long-horizon characteristics of the underlying route.

% \paragraph{Dataset Statistics.}
% We collected over 4,000 navigation trajectories across AirSim urban maps. Among them, 100 trajectories were selected to form the TEST-EASY evaluation benchmark, while the remaining trajectories were used for training. To assess generalization to unseen environments, we further collected 100 navigation trajectories at 14 newly introduced intersections, which constitute the TEST-HARD evaluation benchmark.

\section{Method}
\label{sec:Method}
\subsection{Problem Formulation}

We formalize the language-guided UAV navigation task as a sequential decision-making problem. 
Let $t$ denote the current time step. The agent receives a natural language instruction $I$ and maintains a historical buffer of egocentric visual observations:
\begin{equation}
\mathcal{O}_t = \{ o_i \in \mathbb{R}^{H \times W \times 3} \}_{i=t-L_h+1}^{t},
\end{equation}
where $L_h$ represents the number of historical observation frames and $o_t$ is the current frame from the onboard FPV camera.
Our objective is to employ a unified world model to jointly generate a future action trajectory and predict the corresponding future visual frames. 
Specifically, given $I$ and $\mathcal{O}_t$, the model predicts a future action chunk of length $K$:
\begin{equation}
\mathbf{A}_t = \{ a_j \in \mathbb{R}^8 \}_{j=t}^{t+K-1}.
\end{equation}
Here, each action $a_j$ is represented as an 8-dimensional continuous vector. 
To execute these actions, we apply a deterministic mapping strategy: the continuous vectors are first discretized via a floor operation. If the resulting integer vector corresponds to a valid entry in our pre-defined action table (containing 10 navigation primitives such as ``move forward 3m'' or ``turn left $30^{\circ}$''), the corresponding primitive is executed; otherwise, it executes a ``stop'' action to terminate the navigation task. This design enables the flow matching model to operate in a continuous latent space while ensuring safe and precise discrete maneuvering.
We do not include a backward primitive because the expert A* trajectories in our benchmark are forward-progressing, and reverse motion is typically unsafe under first-person low-altitude navigation with limited rear visibility.
Simultaneously, the model imagines a sequence of future visual observations:
\begin{equation}
\mathbf{V}_t = \{ v_j \in \mathbb{R}^{H \times W \times 3} \}_{j=t+1}^{t+K},
\end{equation}

where each frame $v_j$ represents the predicted visual state resulting from the execution of the corresponding action $a_{j-1}$.
Consequently, the task is formulated as learning a joint generative policy parameterized by $\theta$:
\begin{equation}
p_{\theta}(\mathbf{A}_t, \mathbf{V}_t \mid \mathcal{O}_t, I),
\end{equation}
which captures both future action execution and the spatiotemporal evolution of the scene conditioned on the language instruction.

\subsection{Data Preprocessing}

\paragraph{Language Encoding.}
We employ a text encoder based on T5 with full attention to encode the navigation instruction $I$.
T5 provides context-aware token representations that capture long-range dependencies and fine-grained semantic structure, which is crucial for grounding high-level language instructions into sequential decision-making.
% We set the maximum token length to $L_{txt}=250$ and pad or truncate each instruction accordingly.
We set the maximum token length to $L_{txt}=250$.
The encoded instruction embeddings are denoted as:
\begin{equation}
\mathbf{E}_{I} = \{ e_l \in \mathbb{R}^{D_{txt}} \}_{l=1}^{L_{txt}},
\end{equation}
where $D_{txt}$ is the embedding dimension.

\paragraph{Video Encoding.}
The UAV navigates via discrete motion primitives, where each step induces large spatial displacements and drastic viewpoint changes.
Given these rapid transitions, temporal downsampling would discard critical spatial context required for state estimation.
Therefore, we encode both historical observations and future frames on a per-frame basis to preserve full spatiotemporal fidelity.
We define the full target video sequence $\mathcal{X}$ by concatenating the historical buffer $\mathcal{O}_t$ and the future prediction targets $\mathbf{V}_t$:
\begin{equation}
\begin{aligned}
    \mathcal{X}_t &= \{ x_i \}_{i=t-L_h+1}^{t+K}, &
    x_i &=
    \begin{cases}
        o_i, & t-L_h+1 \le i \le t \quad \text{(History)},\\
        v_i, & t < i \le t+K \quad \text{(Future)}.
    \end{cases}
\end{aligned}
\end{equation}
% \begin{equation}
% \begin{split}
%     \mathcal{X}_t &= \{ x_i \}_{i=t-L_h+1}^{t+K}, \\
%     x_i &=
%     \begin{cases}
%         o_i, & t-L_h+1 \le i \le t \quad \text{(History)},\\
%         v_i, & t < i \le t+K \quad \text{(Future)},
%     \end{cases}
% \end{split}
% \end{equation}
where $o_i$ denotes historical observations and $v_i$ denotes future frames.
We adopt the VAE encoder $E(\cdot)$ from LTX-Video to map each frame $x_i$ into a compressed latent space:
\begin{equation}
z_i = E(x_i), \quad z_i \in \mathbb{R}^{C \times h \times w},
\end{equation}
where $C, h, w$ denote the channel, height, and width of the latent feature map, respectively.
The resulting latent sequence is denoted as $\mathcal{Z}_t = \{ z_i \}_{i=t-L_h+1}^{t+K}$.
Our WorldFly model operates directly in this latent space to predict the future latent states.
During inference, the pixel-space frames are reconstructed frame-wise using the VAE decoder: $\hat{x}_i = D(z_i)$.
This VAE applies aggressive spatial compression while preserving spatial layout, enabling efficient transformer-based modeling.

\subsection{Dual-Branch Coupled Architecture}
We propose a Dual-Branch Coupled Architecture that explicitly separates world modeling and action modeling into two parallel branches:
a \emph{World Model Branch} and an \emph{Action Expert Branch}.
Each branch maintains independent parameters while interacting through periodically inserted dual-branch coupling layers.
Let $g_{\theta_W}$ and $g_{\theta_A}$ denote the networks for the world model and action expert branches, parameterized by $\theta_W$ and $\theta_A$, respectively.

\paragraph{Flow Matching with Aligned Noising.}
In the world-model branch, the historical video latents serve as clean conditioning inputs, while Gaussian noise is added only to the future latents.
Specifically, for each training step we sample a shared timestep $\tau$ and construct noised future latents $\tilde{z}_i(\tau)$. Note that the historical latents $\{z_i\}_{i \le t}$ remain clean and act as conditioning context.
Similarly, for the action expert, we corrupt the future action chunk $\tilde{a}_j(\tau)$ using the same timestep.
By sharing $\tau$, we ensure that the signal-to-noise ratio is consistent across both modalities, preserving the temporal alignment between future dynamics and actions during training.
\paragraph{Training Objective.}
The model is trained to predict the constant velocity vector field that flows from data to noise.
Conditioned on the clean history $\mathcal{O}_t$ and instruction $I$, both branches are optimized via the flow-matching objective.
Here, $\epsilon_i$ and $\eta_j$ denote Gaussian noise sampled for the future latent $z_i$ and action $a_j$, respectively.
The world model loss over future latents is defined as:
\begin{equation}
\mathcal{L}_{W} =
\mathbb{E}_{\tau, \epsilon, \mathcal{Z}}\left[
\sum_{i=t+1}^{t+K}
\left\|
g_{\theta_W}(\tilde{z}_i(\tau), \tau \mid \mathcal{O}_t, I)
-
(\epsilon_i - z_i)
\right\|_2^2
\right].
\end{equation}

\begin{figure*}[t]
  \centering
  \includegraphics[width=\textwidth]{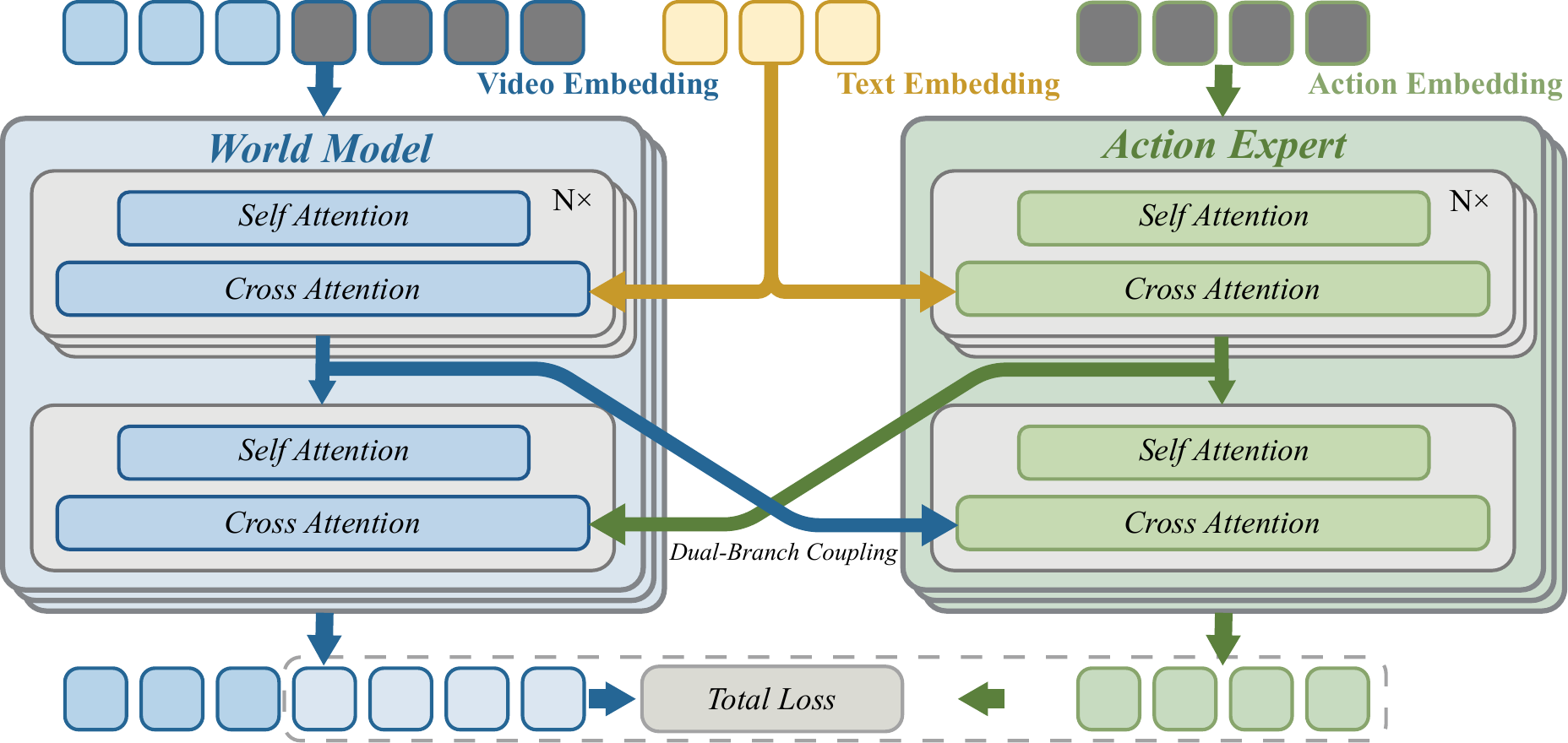}
  \caption{
    Dual-Branch Coupled Architecture. Historical observation frames and language instructions are provided as conditional inputs to a dual-branch flow-matching framework, enabling simultaneous future video latent prediction and action generation. To facilitate cross-modal information exchange, dual-branch coupling blocks are periodically inserted, where cross-attention allows the action expert to leverage imagined future dynamics while the world model is conditioned on planned trajectories.
  }
  \label{fig:worldfly}
\end{figure*}

Correspondingly, the action expert loss over the action chunk is:
\begin{equation}
\mathcal{L}_{A} =
\mathbb{E}_{\tau, \eta, \mathbf{A}}\left[
\sum_{j=t}^{t+K-1}
\left\|
g_{\theta_A}(\tilde{a}_j(\tau), \tau \mid \mathcal{O}_t, I)
-
(\eta_j - a_j)
\right\|_2^2
\right].
\end{equation}
The overall training objective is a weighted sum:
\begin{equation}
\mathcal{L}_{\text{total}} = \mathcal{L}_{W} + \lambda\,\mathcal{L}_{A}.
\end{equation}
In our experiments, we set $\lambda = 1$ to balance the gradients from both branches.

\paragraph{Instruction Attention Block.}
Within each transformer block, the input latent feature $H_{\text{in}}$ first undergoes self-attention, followed by cross-attention with the encoded instruction tokens $\mathbf{E}_I$:
\begin{equation}
H_{\text{out}} = \text{CrossAttn}(\text{SelfAttn}(H_{\text{in}}), \mathbf{E}_I).
\end{equation}
This mechanism ensures that both visual generation and action planning are continuously guided by the high-level navigational goal.

\paragraph{Dual-Branch Coupling Block.}
To enable cross-modal interaction, a Dual-Branch Coupling Block is inserted every $N$ layers.
In this block, the latent features in each branch first undergo self-attention independently. Subsequently, cross-attention is performed, where the self-attended features of one branch act as queries and attend to the key–value representations provided by the other branch. 
Specifically, let $H^W$ and $H^A$ denote the features of the World Model and Action branches, respectively. The coupling is achieved via cross-attention:
% \begin{equation}
% \begin{split}
%     H^W_{\text{out}} &= \text{CrossAttn}(\text{SelfAttn}(H^W_{\text{in}}), H^A_{\text{in}}), \\
%     H^A_{\text{out}} &= \text{CrossAttn}(\text{SelfAttn}(H^A_{\text{in}}), H^W_{\text{in}}).
% \end{split}
% \end{equation}
\begin{equation}
\begin{aligned}
    H^W_{\text{out}} &= \text{CrossAttn}(\text{SelfAttn}(H^W_{\text{in}}), H^A_{\text{in}}), &
    H^A_{\text{out}} &= \text{CrossAttn}(\text{SelfAttn}(H^A_{\text{in}}), H^W_{\text{in}}).
\end{aligned}
\end{equation}
This design allows the action policy to be informed by the imagined future context, while the world model's generation is conditioned on the planned trajectory.

\paragraph{Asymmetric Hidden Dimensions.}
To ensure efficient inference, the action expert branch utilizes a significantly smaller hidden dimension ($D_{act}$) compared to the world model branch ($D_{world}$).
Our coupling mechanism naturally handles this dimensionality mismatch without requiring auxiliary projection layers. 
In the cross-attention mechanism $\text{CrossAttn}(Q, K, V)$, the query $Q$ comes from the target branch, while keys $K$ and values $V$ come from the source branch. 
The learnable weight matrices $W_K$ and $W_V$ in the attention layer implicitly project the source features from their original dimension to the shared attention head dimension.

\section{Experiments}
\label{sec:Experiments}
\subsection{Experimental Setup and Baselines}

We conduct simulation-based evaluations on the \textbf{Urban Canyon Traversal Benchmark}. 
To comprehensively assess model performance, we evaluate on two distinct splits based on the novelty of the intersection layouts:

(1) \textbf{TEST-EASY}: 100 trajectory--instruction pairs sampled from intersections that appeared in the training set. This split tests the model's ability to reproduce expert behaviors within familiar structural configurations.
(2) \textbf{TEST-HARD}: 100 pairs sampled from novel intersections that were strictly excluded from the training data. This split evaluates the model's zero-shot generalization capabilities to unseen spatial geometries and junction types.

\paragraph{Protocol.}
In all experiments, the UAV is initialized at the starting pose of the sampled trajectory. 
At each timestep, the model infers control actions conditioned on the natural language instruction and first-person visual observations. 
Crucially, to ensure a rigorous comparison, the action space is unified across all methods using the discrete OpenFly action primitives. OpenFly and Pi-0-UAV are both fully fine-tuned on the Urban Canyon Traversal benchmark under the same train/test split definition as WorldFly. Each episode is run for at most 100 steps. An episode is considered successful if the final position is within 12 meters of the target; otherwise it is treated as a failure when the maximum episode length is reached or when the agent predicts a stop action while still outside the success threshold. 

Detailed hyperparameters and implementation specifics are provided in the Appendix~\ref{app:extra_exp}.

\begin{table*}[t]
  \caption{Performance comparison across methods on TEST-EASY and TEST-HARD benchmarks.}
  \label{tab:my-wide-table}
  \vskip 0.15in
  \centering
  \begin{small}
    \begin{tabular}{lccc ccc}
      \toprule
      & \multicolumn{3}{c}{TEST-EASY} & \multicolumn{3}{c}{TEST-HARD} \\
      \cmidrule(lr){2-4} \cmidrule(lr){5-7}
      Method & NE (m) & SR (\%) & SPL (\%) & NE (m) & SR (\%) & SPL (\%) \\
      \midrule
      Random        & 46.49 & 2  & 1.79  & 35.30 & 0  & 0     \\
      Pi-0-UAV      & 34.07 & 29 & 27.08 & 41.35 & 10 & 9.43  \\
      OpenFly       & 14.69 & 72 & 58.55 & 35.32 & 16 & 14.92 \\
      % WorldFly (ours) & 3.82 & 95 & 79.51 & 28.76 & 35 & 28.29 \\
      % WorldFly w/o VATV & 7.92  & 87 & 73.25 & 31.08 & 31 & 27.86 \\
      WorldFly (ours) & 7.92  & 87 & 73.25 & 31.08 & 31 & 27.86 \\
      \bottomrule
    \end{tabular}
  \end{small}
  \vskip -0.1in
\end{table*}

\subsection{Experiment Results}
We adopt the four standard evaluation metrics defined in OpenFly~\cite{openfly}:
(i) \textbf{Success Rate (SR)}: The proportion of episodes where the UAV's final position is within 12 meters of the target.
(ii) \textbf{Navigation Error (NE)}: The average Euclidean distance between the final position and the target.
% (3) \textbf{Oracle Success Rate (OSR)}: A metric used to evaluate the UAV's ability to approach the target, which is defined in our experimental evaluation as the success rate of the UAV coming within 20 meters of the target at any point along its trajectory.
(iii) \textbf{Success weighted by Path Length (SPL)}: A rigorous metric balancing success with path efficiency.

The quantitative results are presented in Table~\ref{tab:my-wide-table}.
WorldFly achieves state-of-the-art performance, consistently outperforming all baselines across both splits.

\paragraph{Generalization in Novel Environments.}
Most notably, in the \textbf{TEST-HARD} split (unseen intersections), WorldFly surpasses the strongest baseline (OpenFly) by a significant margin of 15\% in Success Rate (31\% vs. 16\%) and 12.94\% in SPL.
While the absolute success rates decrease for all methods due to the difficulty of zero-shot transfer, WorldFly achieves a performance that is nearly double ($1.9\times$) that of OpenFly.
Additionally, WorldFly reduces the Navigation Error (NE) by approximately 4.2 meters compared to OpenFly (31.08m vs. 35.32m).
This result supports our hypothesis that joint future imagination and action reasoning improves robustness in unseen environments, rather than indicating that reactive policy learning alone is sufficient under large viewpoint transitions.
While reactive baselines suffer from catastrophic performance drops (OpenFly's SR drops from 72\% to 16\%), WorldFly demonstrates significantly stronger robustness.

\paragraph{Advantage of World Modeling.}
Compared to purely reactive VLA models (OpenFly and Pi-0-UAV), WorldFly exhibits drastically lower Navigation Error (NE) in familiar environments.
In the TEST-EASY split, WorldFly reduces the navigation error by nearly 2$\times$ compared to OpenFly (7.92m vs. 14.69m).
We interpret this precision as evidence that the joint imagination-and-action design is beneficial for long-horizon reasoning.
By explicitly imagining future states, WorldFly can anticipate obstacles and turnings in advance, leading to smoother and more proactive maneuvers.
In contrast, reactive models often exhibit higher error rates due to missing turns due to a lack of foresight.

\paragraph{Effectiveness of Dual-Branch Coupling.}
Furthermore, WorldFly significantly outperforms Pi-0-UAV, which also employs a flow-matching backbone but relies on a standard single-stream VLA architecture.
In unseen environments, WorldFly achieves a 3$\times$ higher success rate than Pi-0-UAV (31\% vs. 10\%).
This result is pivotal: it demonstrates that simply applying Flow Matching is insufficient for complex navigation.
Our Dual-Branch Coupled Architecture---which explicitly disentangles world dynamics from action policies while ensuring interaction---is an important contributor to the final performance.
The coupling mechanism allows the action expert to be guided by a dedicated, high-fidelity world simulator, resulting in higher SPL compared to the naive Pi-0-UAV implementation.

\subsection{Ablation Study}
We conduct ablation studies to evaluate the contribution of the Dual-Branch Coupled Architecture (DB), which enables bidirectional interaction between the world model branch and the action expert branch. To ensure a fair comparison, the variant WorldFly w/o DB was trained from scratch using the same dataset, optimization schedule, and hyperparameters as the full WorldFly model.

\begin{table*}[t]
  \caption{Ablation study results on TEST-EASY and TEST-HARD benchmarks.}
  \label{tab:ablation}
  \centering
  \begin{small}
    \begin{tabular}{lccc ccc}
      \toprule
      & \multicolumn{3}{c}{TEST-EASY} 
      & \multicolumn{3}{c}{TEST-HARD} \\
      \cmidrule(lr){2-4} \cmidrule(lr){5-7}

      Method 
      & NE (m) & SR (\%) & SPL (\%) 
      & NE (m) & SR (\%) & SPL (\%) \\

      \midrule
      WorldFly w/o DB 
        & 12.01 & 76 & 61.59 
        & 32.50 & 21 & 18.14 \\

      % WorldFly w/o DB
      %   & 6.92  & 90 & 73.60 
      %   & 29.06 & 23 & 20.04 \\

      WorldFly
        & 7.92  & 87 & 73.25 
        & 31.08 & 31 & 27.86 \\

      % WorldFly
      %   & 3.82  & 95 & 79.51 
      %   & 28.76 & 35 & 28.29 \\

      \bottomrule
    \end{tabular}
  \end{small}
  \vskip -0.1in
\end{table*}

Table~\ref{tab:ablation} presents the results.
Comparing WorldFly with WorldFly w/o DB, the removal of bidirectional coupling leads to significant performance degradation, particularly on the more challenging TEST-HARD set. Specifically, while WorldFly w/o DB achieves a 76\% Success Rate (SR) on TEST-EASY, its performance plummets to 21\% on TEST-HARD (a 55-percentage-point decline). In contrast, the full WorldFly model experiences a more moderate decline from 87\% to 31\% (56 percentage points). This sharper performance decay in the absence of DB indicates that without the Dual-Branch Coupled Architecture, the model struggles to generalize and transfer learned visuomotor associations to difficult environments with larger distribution shifts. Overall, bidirectional coupling is crucial for capturing the intrinsic correlations between visual dynamics (world imagination) and the action space (policy reasoning), thereby enhancing robustness under challenging conditions.

\section{Additional Analysis}
\label{app:add-exp}
\subsection{Inference Latency and Memory.}
\label{app:infer_time}
We measure inference latency on a single NVIDIA A100 GPU using the same 50-step flow-matching denoising schedule as the main evaluation. For a fair measurement of model-side cost, we report the CUDA-event-timed forward latency of the action-only evaluation mode. WorldFly requires approximately 7.81\,s per step, corresponding to a control frequency of approximately 0.5\,Hz. The peak GPU memory is approximately 14.6\,GB allocated. 
% These results indicate that the current system should be viewed as a high-level planner rather than a real-time low-level flight controller. At the same time, the architecture still avoids a separately trained serial world-model-then-policy pipeline and instead jointly optimizes imagination and action generation in a single model.

% \begin{table}[t]
% \centering
% \caption{Inference latency and memory usage on a single NVIDIA A100 GPU under the 50-step flow-matching denoising schedule.}
% \label{tab:latency}
% \begin{tabular}{lcccc}
% \toprule
% Forward (s) & Freq. (Hz) & Peak Mem. (GB) \\
% \midrule
% 7.81 & 0.5 & 14.6 \\
% \bottomrule
% \end{tabular}
% \end{table}

\subsection{Path-Length Analysis.}
We further stratify the evaluation trajectories into short, medium, and long groups based on their path lengths within each split, with each group containing 30 navigation tasks. On TEST-EASY, WorldFly maintains robust performance across all three categories, achieving SR/SPL/NE of 83.3/74.7/7.57 for short, 90.0/73.0/6.76 for medium, and 83.3/68.7/11.10 for long routes. On TEST-HARD, performance exhibits a more pronounced degradation as route length increases: the SR/SPL/NE drops from 47.8/47.0/22.21 (short) to 26.7/23.9/34.66 (medium), and further to 18.8/11.5/35.82 (long). This trend aligns with the inherent challenges of long-horizon generalization. Notably, WorldFly consistently outperforms OpenFly across all categories in TEST-HARD. Even in the most challenging long-route group, where OpenFly fails to complete any tasks (0.0\% SR), WorldFly retains an SR of 18.8\%, demonstrating its superior capability in handling extended navigation trajectories through future-state imagination.

\begin{figure*}[t!]
  \centering
  \includegraphics[width=\textwidth]{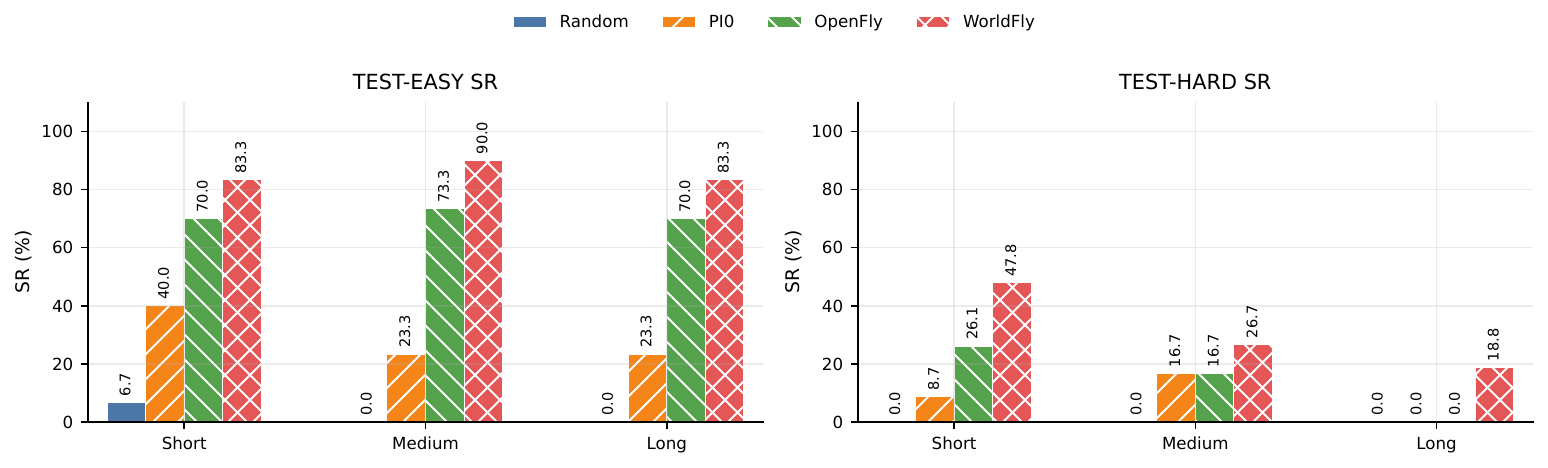}
  \caption{
    Success rates of different methods on short, medium, and long trajectory groups. Trajectories are divided into three groups according to their path lengths. The left figure shows the success rates on TEST-EASY, and the right figure shows the success rates on TEST-HARD.
  }
  \label{fig:path}
\end{figure*}
\begin{table}[t]
\centering
\caption{World-model future video prediction quality (trajectory-level aggregation). Higher is better for PSNR/SSIM, lower is better for LPIPS.}
\label{tab:video_quality}
\begin{tabular}{lccc}
\toprule
Split & PSNR & SSIM & LPIPS \\
\midrule
TEST-EASY & 14.33 & 0.342 & 0.432 \\
TEST-HARD & 13.17 & 0.248 & 0.518 \\
\bottomrule
\end{tabular}
\end{table}
\subsection{World Model Quality.}
To directly quantify the fidelity of imagined future observations, we evaluate the predicted future frames against ground-truth future frames using standard video-prediction metrics. Ground-truth frames are resized to the model's native operating resolution ($224\times224$) before evaluation. As shown in Table~\ref{tab:video_quality}, WorldFly achieves a trajectory-level PSNR/SSIM/LPIPS of 14.33/0.342/0.432 on TEST-EASY and 13.17/0.248/0.518 on TEST-HARD. The consistent degradation from EASY to HARD aligns with the larger distribution shift and more severe viewpoint variation in unseen intersections, providing direct evidence that the world model is non-trivial and produces structurally coherent future predictions even in the harder split.

\section{Conclusion and Limitations}
\label{sec:Conclusion}
In this paper, we introduced \textbf{WorldFly}, a dual-branch coupled Vision-Language-Action model based on world model, designed to address embodied control for UAV in complex low-altitude urban environments.
We proposed a novel architecture that explicitly decouples yet jointly optimizes future video generation and action planning through dense language interaction and periodic modality coupling.
Experimental results on the Urban Canyon Traversal Benchmark demonstrate that WorldFly significantly outperforms strong baselines, including OpenFly and Pi-0-UAV.
These findings empirically validate that high-fidelity generative world modeling serves as a critical catalyst for enhancing the perception and decision-making capabilities of VLA models.
% Future work will focus on improving navigation robustness across diverse downstream tasks and narrowing the sim-to-real gap to deploy this framework in real-world flight missions.
\paragraph{Limitations.}
The main limitation of our work is the unavoidable computational cost brought by future frame prediction. An important future research direction is to apply techniques such as model pruning and distillation to reduce computational costs. 
% We provide detailed experimental data in the appendix~\ref{app:infer_time}.
\bibliographystyle{Ref}  %plainnat,abbrvnat,unsrtnat
% \small
\bibliography{Reference}

@article{openfly,
  title={OpenFly: A Comprehensive Platform for Aerial Vision-Language Navigation},
  author={Gao, Yunpeng and Li, Chenhui and You, Zhongrui and Liu, Junli and Li, Zhen and Chen, Pengan and Chen, Qizhi and Tang, Zhonghan and Wang, Liansheng and Yang, Penghui and others},
  journal={arXiv preprint arXiv:2502.18041},
  year={2025}
}

@article{uav-flow,
  title={UAV-Flow Colosseo: A Real-World Benchmark for Flying-on-a-Word UAV Imitation Learning},
  author={Wang, Xiangyu and Yang, Donglin and Liao, Yue and Zheng, Wenhao and Dai, Bin and Li, Hongsheng and Liu, Si and others},
  journal={arXiv preprint arXiv:2505.15725},
  year={2025}
}

@article{genie-envisioner,
  title={Genie envisioner: A unified world foundation platform for robotic manipulation},
  author={Liao, Yue and Zhou, Pengfei and Huang, Siyuan and Yang, Donglin and Chen, Shengcong and Jiang, Yuxin and Hu, Yue and Cai, Jingbin and Liu, Si and Luo, Jianlan and others},
  journal={arXiv preprint arXiv:2508.05635},
  year={2025}
}

@article{openvla,
  title={Openvla: An open-source vision-language-action model},
  author={Kim, Moo Jin and Pertsch, Karl and Karamcheti, Siddharth and Xiao, Ted and Balakrishna, Ashwin and Nair, Suraj and Rafailov, Rafael and Foster, Ethan and Lam, Grace and Sanketi, Pannag and others},
  journal={arXiv preprint arXiv:2406.09246},
  year={2024}
}

@article{racevla,
  title={Racevla: Vla-based racing drone navigation with human-like behaviour},
  author={Serpiva, Valerii and Lykov, Artem and Myshlyaev, Artyom and Khan, Muhammad Haris and Abdulkarim, Ali Alridha and Sautenkov, Oleg and Tsetserukou, Dzmitry},
  journal={arXiv preprint arXiv:2503.02572},
  year={2025}
}

@article{pi0,
  title={$\pi$0: A vision-language-action flow model for general robot control. CoRR, abs/2410.24164, 2024. doi: 10.48550},
  author={Black, Kevin and Brown, Noah and Driess, Danny and Esmail, Adnan and Equi, Michael and Finn, Chelsea and Fusai, Niccolo and Groom, Lachy and Hausman, Karol and Ichter, Brian and others},
  journal={arXiv preprint ARXIV.2410.24164}
}

@article{videovla,
  title={Videovla: Video generators can be generalizable robot manipulators},
  author={Shen, Yichao and Wei, Fangyun and Du, Zhiying and Liang, Yaobo and Lu, Yan and Yang, Jiaolong and Zheng, Nanning and Guo, Baining},
  journal={arXiv preprint arXiv:2512.06963},
  year={2025}
}

@article{UVA,
  title={Unified video action model},
  author={Li, Shuang and Gao, Yihuai and Sadigh, Dorsa and Song, Shuran},
  journal={arXiv preprint arXiv:2503.00200},
  year={2025}
}

@article{worldvla,
  title={WorldVLA: Towards Autoregressive Action World Model},
  author={Cen, Jun and Yu, Chaohui and Yuan, Hangjie and Jiang, Yuming and Huang, Siteng and Guo, Jiayan and Li, Xin and Song, Yibing and Luo, Hao and Wang, Fan and others},
  journal={arXiv preprint arXiv:2506.21539},
  year={2025}
}

@misc{mind,
      title={MinD: Learning A Dual-System World Model for Real-Time Planning and Implicit Risk Analysis}, 
      author={Xiaowei Chi and Kuangzhi Ge and Jiaming Liu and Siyuan Zhou and Peidong Jia and Zichen He and Yuzhen Liu and Tingguang Li and Lei Han and Sirui Han and Shanghang Zhang and Yike Guo},
      year={2025},
      eprint={2506.18897},
      archivePrefix={arXiv},
      primaryClass={cs.RO},
      url={https://arxiv.org/abs/2506.18897}, 
}

@article{act,
  title={Learning fine-grained bimanual manipulation with low-cost hardware},
  author={Zhao, Tony Z and Kumar, Vikash and Levine, Sergey and Finn, Chelsea},
  journal={arXiv preprint arXiv:2304.13705},
  year={2023}
}

@article{flowmatching,
  title={Flow matching for generative modeling},
  author={Lipman, Yaron and Chen, Ricky TQ and Ben-Hamu, Heli and Nickel, Maximilian and Le, Matt},
  journal={arXiv preprint arXiv:2210.02747},
  year={2022}
}

@inproceedings{bar2025navigation,
  title={Navigation world models},
  author={Bar, Amir and Zhou, Gaoyue and Tran, Danny and Darrell, Trevor and LeCun, Yann},
  booktitle={Proceedings of the Computer Vision and Pattern Recognition Conference},
  pages={15791--15801},
  year={2025}
}

@article{he2025matrix,
  title={Matrix-game 2.0: An open-source real-time and streaming interactive world model},
  author={He, Xianglong and Peng, Chunli and Liu, Zexiang and Wang, Boyang and Zhang, Yifan and Cui, Qi and Kang, Fei and Jiang, Biao and An, Mengyin and Ren, Yangyang and others},
  journal={arXiv preprint arXiv:2508.13009},
  year={2025}
}

@article{mao2025yume,
  title={Yume: An interactive world generation model},
  author={Mao, Xiaofeng and Lin, Shaoheng and Li, Zhen and Li, Chuanhao and Peng, Wenshuo and He, Tong and Pang, Jiangmiao and Chi, Mingmin and Qiao, Yu and Zhang, Kaipeng},
  journal={arXiv preprint arXiv:2507.17744},
  year={2025}
}
% \normalsize

%%%%%%%%%%%%%%%%%%%%%%%%%%%%%%%%%%%%%%%%%%%%%%%%%%%%%%%%%%%%

\appendix
% \begin{figure*}[t]
%     \centering
%     \centerline{\includegraphics[width=\columnwidth]{image/benchmark.pdf}}
%     \caption{
%         Urban Canyon Traversal Benchmark. The top image shows the bird’s-eye-view (BEV) map slices of the AirSim test environment, where shaded regions indicate building clusters. Red points denote intersection locations used for training and the TEST-EASY setting, while green points correspond to intersections used in the TEST-HARD setting. The bottom image illustrates expert UAV navigation trajectories generated by the A* algorithm.
%     }
%     \label{benchmark}
% \end{figure*}
\section{Model Configuration}
\label{app:extra_exp}
% \begin{figure}[ht]
%   \vskip 0.2in
%   \begin{center}
%     \centerline{\includegraphics[width=\columnwidth]{image/comparation.pdf}}
%     \caption{
%        Comparison with baseline methods. We contrast the architectural paradigms of different VLA models. OpenFly is categorized as autoregressive models. In contrast, Pi0 (Pi-0-Fly) represents a flow-matching-based approach, utilizing a continuous generative process via multi-step denoising.
%     }
%     \label{icml-historical}
%   \end{center}
% \end{figure}
\subsection{Training Configuration}

We evaluate three models, \textbf{WorldFly}, \textbf{OpenFly}, and \textbf{Pi-0-UAV}, under identical training settings to ensure a fair comparison. All three models are fully fine-tuned on the Urban Canyon Traversal benchmark using the same train/test split protocol.
% The detailed training configurations are summarized in Table~\ref{tab:training_config}.

We compare WorldFly against the following state-of-the-art baselines:

\begin{itemize}
    \item \textbf{OpenFly}: A 7B parameter model fine-tuned from OpenVLA. 
    To enhance spatiotemporal reasoning, it incorporates sparsely sampled historical visual frames. 
    For a fair comparison with WorldFly, we provide the previous three frames as historical context.

    \item \textbf{Pi-0-UAV}: A flow-matching-based VLA model adapted from the $\pi_0$ architecture, utilizing PaliGemma as the language backbone. 
    It generates continuous actions via multi-step denoising. 
    Consistent with our method, it receives the previous three visual frames as input.
\end{itemize}

\begin{table}[htbp]
\centering
\caption{Training configuration for WorldFly.}
\label{tab:worldfly_config}
\begin{tabular}{lc}
\toprule
Parameter & Value \\
\midrule
Hardware & 4 $\times$ A100 GPUs \\
Training Steps & 40,000 \\
Global Batch Size & 128 \\
Learning Rate & 3e-5 \\
LoRA Enabled & None \\
\bottomrule
\end{tabular}
\end{table}

\begin{table}[htbp]
\centering
\caption{Training configuration for OpenFly.}
\label{tab:openfly_config}
\begin{tabular}{lc}
\toprule
Parameter & Value \\
\midrule
Hardware & 4 $\times$ A100 GPUs \\
Training Steps & 40,000 \\
Global Batch Size & 128 \\
Learning Rate & 5e-4 \\
LoRA Enabled & None \\
\bottomrule
\end{tabular}
\end{table}

\begin{table}[htbp]
\centering
\caption{Training configuration for Pi-0-UAV.}
\label{tab:pi0uav_config}
\begin{tabular}{lc}
\toprule
Parameter & Value \\
\midrule
Hardware & 4 $\times$ A100 GPUs \\
Training Steps & 40,000 \\
Global Batch Size & 128 \\
Learning Rate & 5e-5 \\
LoRA Enabled & None \\
\bottomrule
\end{tabular}
\end{table}

% \begin{table}[htbp]
% \centering
% \caption{Training configuration for different models.}
% \label{tab:training_config}
% \begin{tabular}{lccc}
% \toprule
% Model & Hardware & Training Steps & Global Batch Size \\
% \midrule
% WorldFly & 4 $\times$ A100 GPUs & 40,000 & 128 \\
% OpenFly  & 4 $\times$ A100 GPUs & 40,000 & 128 \\
% Pi-0-UAV      & 4 $\times$ A100 GPUs & 40,000 & 128 \\
% \bottomrule
% \end{tabular}
% \end{table}

All models take three historical frames as input and predict four future frames during training. The action chunk size is set to 4 for \textbf{WorldFly} and \textbf{Pi-0-UAV}, while \textbf{OpenFly} predicts a single action per step (action chunk size = 1).

% \subsection{Inference Configuration}

% \textbf{WorldFly} and \textbf{Pi0} adopt the same flow-matching-based inference strategy with 50 denoising steps. 
% For \textbf{WorldFly}, the coupling period of the dual-branch coefficient is set to $N=5$.

% For the \textbf{WorldFly-Long} variant, a history action chunk buffer of size 6 is maintained, and the weight decay coefficient is set to 0.2.

\subsection{Action Primitives}

We adopt the action primitives defined in \textbf{OpenFly} as the discrete action encoding scheme. Each action primitive represents an atomic low-level motion and is encoded as an 8-dimensional vector. The full set of action primitives is listed in Table~\ref{tab:action_primitives}.

% These primitives are used as a high-level navigation interface for fair comparison with prior UAV VLA benchmarks; a downstream flight controller can execute each primitive while maintaining low-level stabilization. We do not include a backward primitive because the expert routes generated by the A* planner are forward-progressing and backward motion is typically unsafe under first-person low-altitude navigation with limited rear visibility.

\begin{table}[htbp]
\centering
\caption{Definition of action primitives.}
\label{tab:action_primitives}
\begin{tabular}{cll}
\toprule
Index & Action Vector & Description \\
\midrule
0 & [1, 0, 0, 0, 0, 0, 0, 0] & Stop \\
1 & [0, 3, 0, 0, 0, 0, 0, 0] & Move forward 3 m \\
2 & [0, 0, 15, 0, 0, 0, 0, 0] & Turn left 30$^\circ$ \\
3 & [0, 0, 0, 15, 0, 0, 0, 0] & Turn right 30$^\circ$ \\
4 & [0, 0, 0, 0, 2, 0, 0, 0] & Move up 3 m \\
5 & [0, 0, 0, 0, 0, 2, 0, 0] & Move down 3 m \\
6 & [0, 0, 0, 0, 0, 0, 5, 0] & Move left 3 m \\
7 & [0, 0, 0, 0, 0, 0, 0, 5] & Move right 3 m \\
8 & [0, 6, 0, 0, 0, 0, 0, 0] & Move forward 6 m \\
9 & [0, 9, 0, 0, 0, 0, 0, 0] & Move forward 9 m \\
\bottomrule
\end{tabular}
\end{table}

\subsection{Additional Experimental Details.}
We include a \textbf{Random} baseline for reference. Specifically, \textbf{Random} denotes a policy that samples actions uniformly at random from the same discrete OpenFly action primitive set used by all methods. At each environment step, the agent independently draws one action from the \textbf{ten} available primitives and executes it, without conditioning on the instruction or visual observations.

\textbf{WorldFly} and \textbf{Pi-0-UAV} both adopt flow matching and perform the same 50-step denoising procedure to obtain the predicted action chunk; additionally, \textbf{WorldFly} predicts future video latents during the same denoising process. In contrast, \textbf{OpenFly} is an autoregressive model that predicts actions in an autoregressive manner. For all three methods, at each environment step we execute the \emph{first} action of the predicted action chunk to interact with the environment.

At evaluation time, each episode is run for at most 100 steps. An episode is marked successful if the final UAV position is within 12 meters of the target. Otherwise, it is treated as a failure when the maximum episode length is reached or when the model predicts a stop action while still outside the success threshold. Navigation instructions are generated with an LLM-based instruction constructor built on the OpenFly toolchain using Qwen3-VL.

\section{Model Hyperparameters}
As shown in Table~\ref{tab:hyperparams}, we report the relevant hyperparameters of our model. 
It is worth noting that the world model branch and the action expert branch adopt different hidden dimensions. 
All ablation experiments are trained using the same configuration.

\begin{table}[htbp]
\centering
\caption{Model hyperparameters.}
\label{tab:hyperparams}
\begin{tabular}{lp{8cm}l}
\toprule
\textbf{Parameter} & \textbf{Description} & \textbf{Value} \\
\midrule
\textit{Data Processing} & & \\
$H, W$ & Input RGB frame resolution (height $\times$ width) & $224 \times 224$ \\
$L_h$ & Length of historical observation buffer (frames) & $3$ \\
$K$ & Length of predicted future action/frame chunk & $4$ \\
$L_{txt}$ & Maximum token length for instruction encoding (T5) & $250$ \\
$D_{txt}$ & T5 text embedding dimension & $4096$ \\

\midrule
\textit{WorldFly Architecture} & & \\
$D_{world}$ & Hidden dimension of World Model branch & $2048$ \\
$D_{act}$ & Hidden dimension of Action Expert branch & $512$ \\
$N_{layer}$ & Number of model layers & $28$ \\
$N$ & Frequency of Dual-Branch Coupling Block insertion & $5$ \\
\midrule
\textit{WorldFly Inference} & & \\
$m$ & Size of Action Cache & $4$ \\
$\alpha$ & Decay rate for cached action sequence voting weights & $0.2$ \\
\bottomrule
\end{tabular}
\end{table}

\section{Joint Video and Action Prediction Examples}

\begin{figure*}[htbp]
  \centering
  \includegraphics[width=\textwidth]{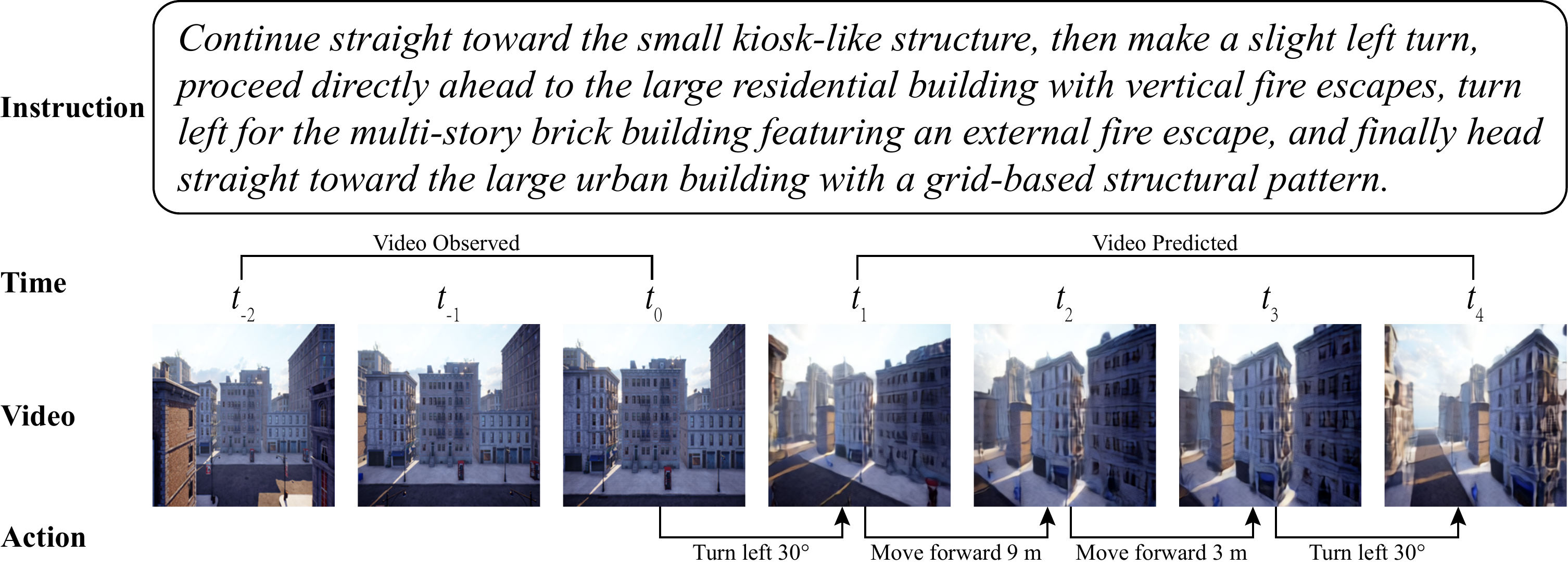}
  \caption{
    Joint video generation and action prediction with WorldFly.
  }
  \label{fig:jointPre}
\end{figure*}

As shown in Figure~\ref{fig:jointPre}, we visualize a joint inference example of \textbf{WorldFly} for future video generation and action prediction.
The input consists of 3 frames of historical observations and the natural language instruction; flow-matching denoising is performed for 50 steps, and 4 future frames are predicted.
This case is selected from the \textbf{TEST-EASY} split to qualitatively demonstrate WorldFly's capability of simultaneously performing future imagination and action execution.
For visualization, the imagined future video is obtained by decoding the predicted future latents frame by frame using \textbf{LTX-Video VAE} .

%%%%%%%%%%%%%%%%%%%%%%%%%%%%%%%%%%%%%%%%%%%%%%%%%%%%%%%%%%%%

\newpage

\end{document}